\documentclass[conference]{IEEEtran}
\usepackage{comment}
\usepackage{multirow}
\usepackage{graphicx}
\usepackage{pifont}
\usepackage{wrapfig}
\usepackage{subcaption}
\usepackage{amsmath,amsfonts,amssymb, bm}
\usepackage{xcolor}
\usepackage{booktabs,siunitx}
\definecolor{DodgerBlue}{RGB}{30,144,255}
\usepackage{todonotes}

\newcommand{\SO}{\mathrm{SO}}
\makeatletter
\let\NAT@parse\undefined
\makeatother
\usepackage[colorlinks=true]{hyperref}

\IEEEoverridecommandlockouts                              

\hypersetup{breaklinks=true,bookmarks=true,colorlinks=true,linkcolor=DodgerBlue,urlcolor=DodgerBlue, citecolor=DodgerBlue,}
\title{\LARGE \bf
Residual Rotation Correction using Tactile Equivariance
}

\author{Yizhe Zhu$^{1}$, Zhang Ye$^{1}$, Boce Hu$^{1}$, Haibo Zhao$^{1}$, Yu Qi$^{1}$, Dian Wang$^{2\dagger}$, and Robert Platt$^{1\dagger}$

\thanks{$^{\dagger}$Equal Advising. Corresponding to \tt\small dianwang@stanford.edu }\

\thanks{$^{1}$Yizhe Zhu, Zhang Ye, Boce Hu, Haibo Zhao, Yu Qi and Robert Platt are with Northeastern University, Massachusetts, MA, USA {\tt\small \{zhu.yizhe, ye.zhang1, hu.boce, zhao.haib,  qi.yu2, r.platt@northeastern.edu }}%
\thanks{$^{2}$Dian Wang is with Stanford University, California, CA, USA
        {\tt\small dianwang@stanford.edu }}%
}

\begin{document}

\maketitle
\thispagestyle{empty}
\pagestyle{empty}

\begin{abstract}
Visuotactile policy learning augments vision-only policies with tactile input, facilitating contact-rich manipulation. However, the high cost of tactile data collection makes sample efficiency the key requirement for developing visuotactile policies. We present EquiTac, a framework that exploits the inherent $\SO(2)$ symmetry of in-hand object rotation to improve sample efficiency and generalization for visuotactile policy learning. EquiTac first reconstructs surface normals from raw RGB inputs of vision-based tactile sensors, so rotations of the normal vector field correspond to in-hand object rotations. An $\SO(2)$-equivariant network then predicts a residual rotation action that augments a base visuomotor policy at test time, enabling real-time rotation correction without additional reorientation demonstrations. On a real robot, EquiTac accurately achieves robust zero-shot generalization to unseen in-hand orientations with very few training samples, where baselines fail even with more training data. To our knowledge, this is the first tactile learning method to explicitly encode tactile equivariance for policy learning, yielding a lightweight, symmetry-aware module that improves reliability in contact-rich tasks.
\href{https://yizhezhu0925.github.io/equitac.github.io/}{https://yizhezhu0925.github.io/equitac.github.io/}.

\end{abstract}

\section{Introduction}

As robots increasingly rely on touch to perform precise, contact-rich interactions, developing visuotactile policies has become a key challenge in robotic manipulation. While data-driven visuomotor policy learning has achieved remarkable progress, extending these advances to tactile sensing remains fundamentally data-constrained. Unlike visual data, tactile signals are harder to scale, since collecting sufficient tactile interactions to cover diverse contact conditions is extremely expensive, and publicly available tactile datasets are still limited in both scale and diversity~\cite{yang2022touch, huang2024normalflow, fu2024touch}. Thus, learning tactile manipulation policies, or even tactile reasoning modules that augment vision-based policies, demands far greater sample efficiency than comparable vision-only pipelines. 

To mitigate the above data constraints, equivariant learning, i.e., embedding task symmetries as inductive bias in neural networks, offers a direct route to improving sample efficiency and generalization. By baking in spatial regularities that would otherwise need to be learned from data, equivariant policy learning in robotic manipulation has consistently shown gains across imitation~\cite{yang2024equibot,zhang2025canonical} and RL~\cite{wang2022mathrm} pipelines, enabling few-shot and on-robot learning~\cite{zhao2025hierarchical, zhu2022sample}. However, existing studies remain largely vision-centric: they design equivariance for image or scene geometry, while equivariance in tactile learning remains under-explored. In particular, prior equivariant robot-learning approaches do not explicitly model the object-in-gripper rotational symmetry that underlies tactile feedback, which is crucial for contact-rich manipulation.

\begin{figure}[t]
    \centering
\includegraphics[width=0.5\textwidth]{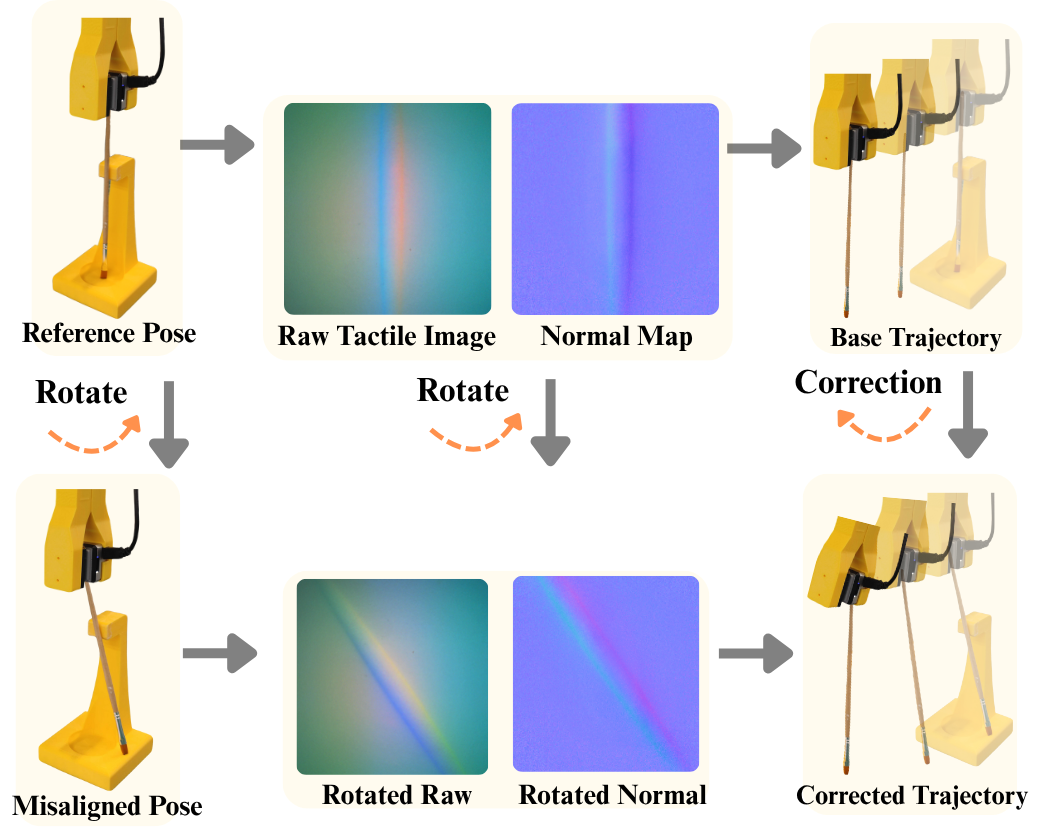}
    \caption{\textbf{Equivariance in EquiTac.} When the tactile observation is rotated, the predicted action rotates consistently.}
    \label{fig:1}
\end{figure}

To close this gap, this paper introduces \emph{EquiTac}, the first equivariant tactile learning pipeline. Our core idea is to leverage in-hand object orientation symmetry: when the grasped object rotates within the gripper, the tactile signal should transform predictably. However, in vision-based tactile sensors, the raw RGB intensities are often distorted by internal illumination effects (e.g., RGB LEDs), making naive in-plane image rotation inconsistent with the true contact geometry. We address this by reconstructing surface normal maps from tactile images and decomposing them into equivariant (in-plane $x,y$ vector) and invariant (out-of-plane normal-$z$) components. This representation restores rotation-consistent behavior and provides the correct carriers for $\SO(2)$-equivariant reasoning. Building on this representation, we propose a residual policy correction framework, where a rotation residual predicted by an equivariant network augments a trained manipulation policy at test time to correct the gripper's orientation. \autoref{fig:1} illustrates the core idea of this equivariant correction: when the object rotates in hand, the equivariant network observes a rotated normal map and predicts a corresponding target gripper rotation, from which an angular residual is computed to correct the base policy. By injecting this symmetry-aware tactile correction, the policy generalizes zero-shot to unseen in-hand object orientations without retraining.

Concretely, we make the following contributions:
\begin{enumerate}
    \item We propose EquiTac, which encodes in-hand rotational symmetry via normal-map parameterization and an $\SO(2)$-equivariant network to support symmetry-consistent tactile representations. To the best of our knowledge, this is the first work to introduce equivariance into tactile learning.
    \item We develop an equivariant tactile residual correction pipeline, a lightweight module that predicts a target gripper rotation from a single tactile frame and computes an angular residual to correct an existing policy zero-shot for unseen in-hand orientations.
    \item We empirically validate EquiTac on in-hand angle estimation and residual policy correction, demonstrating higher prediction accuracy and strong zero-shot generalization to unseen object orientations.
\end{enumerate}

\section{related work}

\textbf{Visuotactile Manipulation.} 
Tactile feedback is critical for contact-rich manipulation, especially when vision alone cannot reliably determine an object’s in-hand orientation~\cite{andrychowicz2020learning, khandate2023sampling, Suresh2021TactileSLAM, Sodhi2021TactileFactorGraph, Villalonga2021FirstTouch,Kelestemur2022TactileAAMAS}. Recent visual–tactile imitation-learning methods fuse modalities to reconstruct object pose~\cite{huang20243d, nonnengiesser2025hand,huang2025vt}, but they generally do not adjust policy outputs online using contact information. These approaches achieve promising results, but they generally lack the ability to actively adjust policy outputs in real time using contact information. Some methods incorporate shear-force sensing and deploy auxiliary models running at different inference rates to refine the original policy’s actions online~\cite{xue2025reactive}. However, they do not use the object’s rotation information to update the policy outputs in real time. Other studies designed an “orientation readjustment” phase during data collection to satisfy downstream requirements~\cite{zhu2025touch}. However,  such task-specific adjustment strategies are not broadly generalizable and substantially increase data collection costs. In contrast, our method preserves the original vision-based policy structure while making real-time, contact-driven adjustments without adding an extra, manually crafted orientation-adjustment stage. 

\textbf{Residual learning in robotics.} 
Residual learning has proven effective in robotics. Methods can be grouped into three categories: reinforcement learning; human corrections; and interactive imitation learning. RL-based work~\cite{johannink2019residual,ankile2025imitation}, such as Policy Decorator~\cite{yuan2024policy}, EXPO~\cite{dong2025expo} and ResiP~\cite{ankile2025imitation}, learns residual corrections through interaction with off-policy algorithms or by combining behavioral cloning with RL. Methods involving human corrections, such as TRANSIC~\cite{jiang2024transic} and CR-DAgger~\cite{xu2025compliant}, collect teleoperation data or force-aware adjustments from operators and use this to train residuals with supervision. Interactive imitation learning, as demonstrated by HG-DAgger~\cite{kelly2019hg}, enables experts to intervene during execution and provide corrective demonstrations. However, these approaches often require thousands of online interactions, substantial human supervision during deployment, and complex sim-to-real transfer. Furthermore, the residual networks are often similar in size to the base policy. Our method takes a different approach. We encode rotational equivariance from tactile data to achieve zero-shot generalization, eliminating the need for online RL or sim-to-real transfer. The correction module can be trained with supervision on a single example. This reduces sample complexity and system overhead, enabling direct deployment in a real environment.

\textbf{Equivariance in Robotics.}
Equivariance has been shown to boost performance and improve sample efficiency~\cite{wang2022mathrm, zhu2022sample,huang2022equivariant, hu2024orbitgrasp, zhao2025hierarchical, huang2024fourier, qi2025two, hu2025push, zhu2025equact, wang2023onrobot, wang2022equiq, zhu2023grasp}, allowing policies to learn effectively from far fewer demonstrations; it has been applied across open-loop and closed-loop settings as well as diverse dataset generation. Recently, the idea has been extended to closed-loop diffusion policies: Equivariant Diffusion Policy~\cite{wang2024equivariant} augments diffusion policies with an SO(2)-equivariant architecture to leverage task symmetries, yielding better generalization and data efficiency. Beyond planar symmetries, 3D-Spherical Projection~\cite{hu20253d} achieves SO(3) equivariance from a single RGB camera by projecting features onto a sphere for real-time visuomotor control. However, these approaches typically do not explicitly model object-in-gripper rotation equivariance, which is crucial for contact-rich or fine manipulation tasks. To address this gap, we develop a new equivariant framework that enables real-time, visuo-tactile action correction using tactile and visual sensing.

\section{Background}
\begin{figure*}[t]
    \vspace{0cm}
    \begin{center}
    \includegraphics[width=1\textwidth]{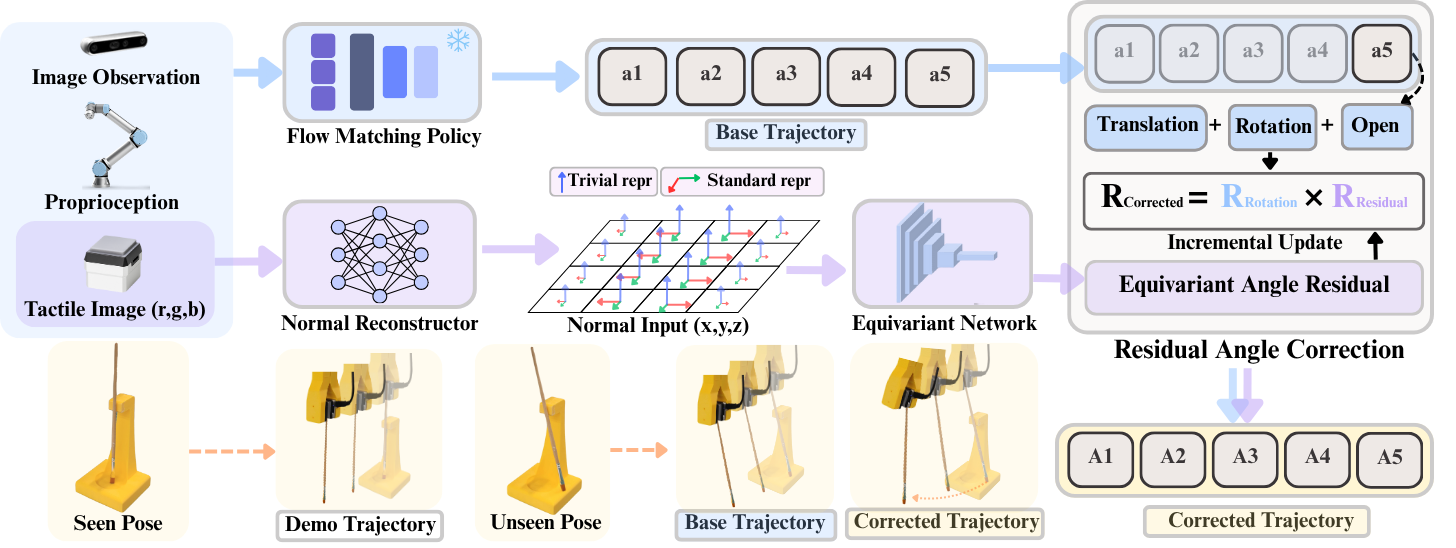}
    \captionof{figure}{\textbf{Overview of our tactile-guided manipulation framework with equivariant orientation correction.
    } The system begins with a Flow Matching Policy (top) that predicts basic action chunks from multimodal inputs including robot proprioception, tactile images, and three camera views. During action execution (middle), tactile images are processed through a Normal Reconstructor to obtain normal maps, which are fed into an $\SO(2)$-equivariant network. The equivariant network will predict the angular residual between the object's current and target orientations, enabling real-time correction of the action chunk to correct for misalignment. The bottom row shows (i) the data-collection setup with ideal object placement and (ii) the results of executing the base and corrected trajectories under placement deviations at rollout.}
    \label{fig:teaser}
    \end{center}
\end{figure*}
\subsection{Equivariance}
Let $G$ be a symmetry group, a function $f:\mathbb{R}^n\!\to\!\mathbb{R}^m$ is \emph{equivariant} with respect to $G$ if applying a group transform before $f$ is equivalent to transforming the output after $f$:
\[
f\!\big(\rho_{\text{in}}(g)\,x\big)=\rho_{\text{out}}(g)\,f(x),\qquad \forall g\in G.
\]

In this work, we focus on planar rotations $G = \SO(2)$. Here, $\rho_\text{in}$ and $\rho_\text{out}$ are called the group representations that define how the input and output  transform under a rotation $g\in \mathrm{SO}(2)$. For example, $\rho=1$ corresponds to the trivial (invariant) representation acting on scalars under rotation, and $\rho=R_g=\begin{bmatrix}
\cos g & -\sin g\\
\sin g & \cos g
\end{bmatrix}$ represents the standard represnetations of $\SO(2)$ acting on vectors.

\subsection{Flow Matching Policy}\label{Flow_matching}
Flow-based policy learning \cite{black2024pi_0, zhang2024affordance} is a class of imitation learning methods that model action generation as a continuous-time transport process using learned velocity fields. These methods learn to transform noise samples into structured actions through ordinary differential equation (ODE) integration, conditioned on multi-modal observations. Formally, given an observation encoding $c$ (which may include visual features, tactile signals, and proprioceptive states) and a flow time $t \in [0,1]$, the policy learns a time-dependent velocity field $v_{\theta}(x, t, c)$ that transports a noise sample $x_0 \!\sim\! \mathcal{N}(0, \sigma^2 I)$ toward a target action $x_1$. During training, the method constructs interpolated states via $x_t = (1 - t)\,x_0 + t\,x_1$, where the ground-truth velocity is $\frac{d x_t}{d t} = x_1 - x_0$. The model is trained to minimize the velocity-matching objective:
\[
\mathcal{L} = \mathbb{E}_{x_0,x_1,t} \big\|\,v_{\theta}(x_t, t, c) - (x_1 - x_0)\,\big\|^2.
\]
At test time, the policy generates actions by sampling $x_0 \!\sim\! \mathcal{N}(0, \sigma^2 I)$ and numerically integrating the learned ODE $\frac{d x}{d t} = v_{\theta}(x, t, c)$ with initial condition $x(0) = x_0$. The integration is performed from $t=0$ to $t=1$ using $N$ Euler steps: $x_{i+1} = x_i + \frac{1}{N}\, v_{\theta}(x_i, t_i, c)$ where $t_i = \frac{i}{N}$, yielding the final action $x_N$.
\section{Method}
\subsection{Problem Statement}

We study closed-loop visuotactile policy learning for contact-rich robotic manipulation that can generalize to unseen object orientations, formulated as imitation learning from expert demonstrations. Specifically, we consider an expert demonstration dataset $\mathcal{D}=\{(\mathcal{O},\mathcal{A})\}_{t=1}^{N}$, where $\mathcal{O}=\mathcal{O}_v\times \mathcal{O}_p \times \mathcal{O}_\text{tac} $ includes visual observations from RGB cameras, proprioceptive readings from the robot, and tactile measurements from a touch sensor. $\{a_1, a_2, \dots, a_m\}\in \mathcal{A}$ denotes an action chunk consisting of a sequence of robot actions in the next $m$ time steps. 

The goal is to learn a policy that integrates a visuomotor base policy $\pi_b$ with a tactile residual policy $\pi_r$. The base policy $\pi_b: \mathcal{O} \to \mathcal{A}$ first predicts an action chunk from multi-modal observations, then the tactile residual policy $\pi_r: \mathcal{O}^{\mathrm{tac}} \times \mathcal{A} \to \mathcal{A}$ generates the corrected action chunk from the tactile feedback and the base actions. Together, the goal for the composed policy $\pi = \pi_r \circ \pi_b$ is to achieve robust manipulation under varying in-hand object orientations. 
\subsection{Overview of EquiTac}

\autoref{fig:teaser} shows the overview of EquiTac. Given a multimodal observation consisting of RGB camera images, robot proprioception, and a tactile image, we first use a flow-matching policy $\pi_b$ to produce a base action chunk. In parallel, an equivariant tactile residual policy $\pi_r$ will generate an in-hand rotation residual, which is applied to the base action chunk to correct the angular error caused by the unseen in-hand object pose.

\begin{wrapfigure}[14]{r}{0.15\textwidth}  
    \centering
    \vspace{-8pt}     \includegraphics[width=\linewidth]{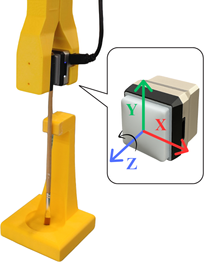}
    \caption{Sensor on the gripper fingertip, $z$-axis denotes the finger normal.}
    \label{fig:Frame}
\end{wrapfigure}

Our equivariant tactile residual policy $\pi_r$ has three main steps. First, we use a learnable mapping to convert the raw tactile image into a normal map. Second, taking the normal map as the input, an equivariant network predicts a target \emph{in-hand yaw} rotation. Let $\{x, y, z\}$ be the fingertip local frame with $z$ the contact normal, in-hand yaw is the rotation about $z$, i.e., the rotation about the fingertip surface normal (\autoref{fig:Frame}). Finally, we calculate a rotation residual between the predicted target yaw and the current gripper orientation, and apply it incrementally to the base action sequence to refine the gripper’s rotation.

\subsection{Surface Normal Map Reconstruction}
To enable rotational equivariance, the tactile representation must transform consistently with the object’s true in-hand rotation. Vision-based tactile sensors illuminated by fixed RGB LEDs do not satisfy this property directly, as identical local geometry can yield different colors as the object rotates. Consequently, naively using RGB image rotation as the representation does not align with contact geometry.

\begin{figure}[t]
    \centering
\includegraphics[width=0.47\textwidth]{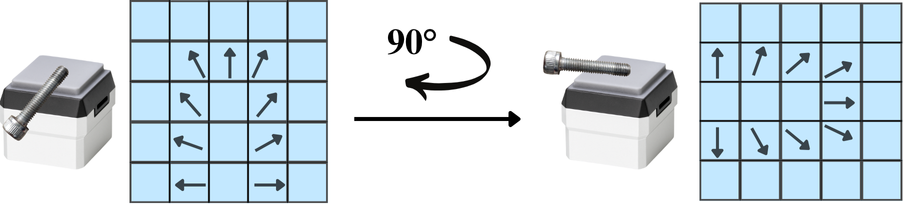}
    \caption{\textbf{Equivariance of the normal map.} When the object rotates in hand, the normal map co-rotates as a vector field.}
    \label{fig:Equi}
\end{figure}

We resolve this by reconstructing a surface normal map $N\in\mathbb{R}^{3\times H\times W}$ using a lightweight MLP, where each pixel of the normal map stores a unit normal vector,
$\vec{n} = ({n_x}, {n_y}, {n_z})$. Under an in-hand object rotation $g\in\mathrm{SO}(2)$ about the fingertip normal, all the pixels in the normal map will rotate accordingly. Moreover, for each individual vector $\vec{n}$, the in-plane components $(n_x, n_y)$ rotate as a 2-vector, while $n_z$ remains invariant,
\[
g\cdot\vec{n} = \begin{bmatrix}
R_g & 0 \\
0 & 1
\end{bmatrix} (n_x, n_y, n_z)^T,
\]
where $R_g$ is the $2\times 2$ rotation matrix for $g$.
\autoref{fig:Equi} shows an example of such transformation. We denote $g\cdot N$ as this vector field rotation. This representation provides the correct carriers for $\mathrm{SO}(2)$-equivariant reasoning and ensures the representation co-rotates with the object.

We follow the standard GelSight calibration procedure~\cite{wang2021gelsight,huang2024normalflow} to obtain pixel-wise ground truth geometry for training the normal reconstructor.
Specifically, a metal calibration sphere is pressed against the sensor surface at multiple locations to generate contact samples with known geometry. For each contact, the local surface gradient is computed analytically from the known spherical geometry.
Using these gradients as supervision, an MLP is trained to directly predict the surface gradient at each pixel from its color and spatial location $(R, G, B, U, V)$. The predicted gradient map is then converted into a surface normal map via standard gradient-to-normal conversion.

\subsection{Equivariant Angular Residual Tracking} \label{Equivariant Angular Residual Tracking}

After acquiring the normal map $N$, we aim to train an equivariant network $\phi: N\mapsto (\cos-\alpha_t, \sin-\alpha_t)$ that predicts the target in-hand yaw $\alpha_t$ in the form of a unit vector on a unit circle. The model satisfies 
\[
\phi (g\cdot N) = g\cdot \phi(N) = R_g (\cos-\alpha_t, \sin -\alpha_t)^T,
\]
where $R_g$ is the standard $2\times 2$ rotation matrix. Thus, when the object rotates on the finger by $g\in \SO(2)$, the input normal map $N$ becomes $g\cdot N$, and the output target in-hand yaw vector will counter-rotate accordingly, compensating for the in-hand object rotation. This property facilitates zero-shot generalization, achieving precise angular estimation without training on tactile images with different object orientations. 

In practice, although an ideal equivariant function can theoretically generalize to unseen orientations without additional training, the discretization of $\SO(2)$ in implementation leads to incomplete coverage of the rotation space.

To address this, we augment the training data by randomly rotating the entire normal map with uniformly sampled angles, thereby preserving geometric consistency with the equivariant structure of the model.

\subsection{Equivariant Angular Residual Correction}

We use Flow Matching (\ref{Flow_matching}) as the base policy $\pi_b$ to progressively denoise an initial noisy action and obtain a base action chunk $\{a_i=(T^{b}_i,p_i)\}_{i=1}^m$, where $T^{b}_i\in \mathbb{R}^{4\times 4}$ is a transformation matrix representing the center of the gripper fingertips in the world frame, and $p_i$ is the gripper command. The equivariant tactile residual policy $\pi_r$ uses $\phi$ to estimate the in-hand yaw target $\alpha_t$, then apply a proportional update to $T^b_i$ by a rotation in the gripper frame. Specifically, given the current in-hand yaw $\alpha$ we define the residual transformation matrix as a rotation about the z-axis in the gripper frame (\autoref{fig:Frame}),
\[
T^r = \begin{bmatrix}
\cos K_p(\alpha_t-\alpha) & -\sin K_p(\alpha_t-\alpha) & 0 & 0 \\
\sin K_p(\alpha_t-\alpha) & \cos K_p(\alpha_t-\alpha) & 0 & 0 \\
0 & 0 & 1 & 0\\
0 & 0 & 0 & 1
\end{bmatrix},
\]
where $K_p \in (0,1)$ is a proportional gain. The residual policy correction is achieved by right multiplication $T^b_i\cdot T^r$. This proportional update mechanism ensures that each value in the angular residual gradually adjusts the rotation angle in the current action chunk toward the target. In the actual implementation, our overall policy runs at two rates: the base Flow Matching policy emits action chunks at a lower rate, while a lightweight equivariant module provides high-frequency in-hand yaw corrections.

\subsection{Implementation Details}
Our network architecture follows the theoretical formulation of $\mathrm{SO}(2)$ equivariance described in \ref{Equivariant Angular Residual Tracking}, which is a four-layer $C_8$-equivariant convolutional neural network, implemented using the ESCNN~\cite{cesa2022program} library. We adopt a double-angle representation on the output, where the network is trained to predict $(\cos -2\alpha_t, \sin -2\alpha_t)$. At inference time, the orientation is recovered via $\alpha_t = -\frac{1}{2} \arctan2(\sin -2\alpha_t, \cos -2\alpha_t)$. This double-angle representation has two advantages. First, it maps both $\alpha_t$ and $\alpha_t + \pi$ to the same output vector, preventing the ambiguity caused by many objects that are symmetric over $\pi$ rotation. Second, it ensures $\alpha_t\in [-\frac{\pi}{2}, \frac{\pi}{2} )$, restricting corrections to the forward-facing semicircle and preventing inward-pointing actions that could cause collisions.

\section{Model Evaluations}
\begin{figure}[t]
\centering
\includegraphics[width=1\linewidth]{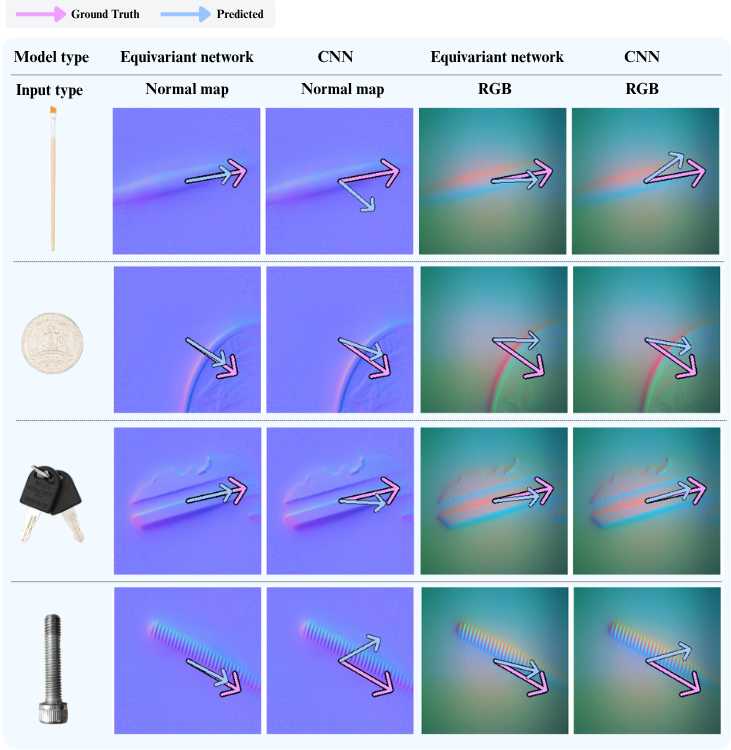}
\caption{Qualitative comparison of angular estimation across different model configurations. }
\label{fig:compare_visual}
\end{figure}

\subsection{Ablation on Model Design and Input Representation} 
We conduct ablation studies to evaluate the impact of the equivariant network $\phi$ and input representation on in-hand object orientation prediction. Experiments are performed on four representative objects: \emph{Brush}, \emph{Coin}, \emph{Key}, and \emph{Screw}. For each object, $\phi$ is trained on a single object orientation with data augmentation and tested on 100 randomly sampled orientations, which are kept consistent across all variants for fair comparison. The full model is compared against three modified versions.
\begin{itemize} 
\item \textbf{No Normal Map}: The model takes raw tactile RGB images as input instead of reconstructed surface normal maps.
\item \textbf{No Equivariance}: All $\SO(2)$-equivariant layers are replaced with standard conv layers, while keeping the same backbone architecture and data augmentation.

\item \textbf{No Data Augmentation}: The model is trained without applying any rotational data augmentation techniques.
\end{itemize}

\begin{table}[t]
\centering
\caption{\textbf{Ablation study on model design and input representation.} Angular estimation errors (in degrees) across four objects, with each row showing a variant of the model with one component removed.}
\setlength{\tabcolsep}{6.5pt}
\footnotesize
\begin{tabular}{l *{4}{c}|c}
\toprule
Configuration & Brush & Coin & Keys & Screw & Mean \\
\midrule
Full Model (Ours) & \textbf{2.7} & \textbf{2.4} & \textbf{3.0} & \textbf{3.7} & \textbf{2.9} \\
\midrule
No Normal Map & 29.2 & 26.3 & 30.4 & 23.8 & 27.4 \\
No Equivariance & 10.5 & 19.6 & 13.3 & 10.5 & 13.5 \\

No Data Augmentation & 9.3 & 12.5 & 7.6 & 8.3 & 9.4 \\
\bottomrule
\end{tabular}
\label{tab:ablation}
\end{table}

As shown in \autoref{tab:ablation}, the full model with all components achieves a mean angular estimation error of only 2.9$^\circ$, indicating its high precision. Qualitative results across different configurations are visualized in \autoref{fig:compare_visual}, highlighting the improvement in angular alignment achieved by our full model. Replacing the normal map with raw RGB tactile inputs causes the most significant degradation, increasing the mean error to approximately 27$^\circ$. This highlights that normal-map representations are essential for capturing geometry-consistent rotation cues and accurately reflecting object orientation. Removing $\SO(2)$-equivariance in the network while retaining the normal map increases the mean error to about 14$^\circ$, confirming that the equivariant structure further improves angular estimation accuracy. 
Finally, to assess the role of data augmentation, we train the equivariant network without rotation augmentation. This yields an error of 9.4$^\circ$ on average, which is slightly higher than the full model but still lower than the non-equivariant baseline trained with augmentation.
This result suggests that while data augmentation provides a benefit, the primary performance gains arise from the model’s equivariant architecture and the use of normal-map inputs.

\subsection{Robustness to Input Variations} 
\begin{figure}[t]
\centering

\centering
\includegraphics[width=0.48\textwidth]{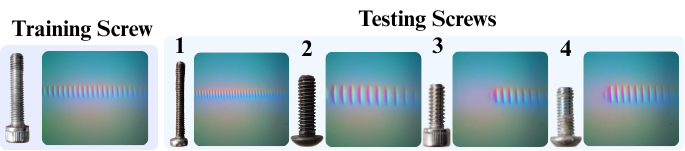}
\caption{\textbf{Screw geometries for generalization evaluation.} Left: training screw. Right: four novel test screws with diverse thread patterns and textures.}
\label{fig:screw_compare}
\end{figure}

We evaluate the generalization capability of our model by training the network $\phi$ on tactile data from a single screw instance and testing it on several unseen screws with similar but distinct geometries (\autoref{fig:screw_compare}). Each test screw has a different thread pattern, and the model is evaluated on 100 randomly sampled object orientations per instance.

As shown in \autoref{tab:screw_generalization}, the model accurately estimates the angular residuals for these unseen screw types. Despite being trained on a single screw geometry, it maintains low prediction errors on novel instances, with only a modest increase compared to the training instance. This demonstrates a strong generalizability of our model, attributed to the equivariant structure and geometry-aware normal map representation.

\vspace{0.5em} 

\begin{table}
\centering
\captionof{table}{\textbf{Generalization to unseen screw geometries.} Angular errors (in degrees) when trained on a single screw type and tested on four novel thread patterns.}
\vspace{0.2em}
\small
\setlength{\tabcolsep}{3pt}
\renewcommand{\arraystretch}{1.05}
\begin{tabular*}{0.98\linewidth}{@{\extracolsep{\fill}} l c c c c c @{}}
\toprule
Method & Training & Test 1 & Test 2 & Test 3 & Test 4 \\
\midrule
Ours & 3.7 & 4.6 & 5.3 & 4.1 & 5.6 \\
\bottomrule
\end{tabular*}
\label{tab:screw_generalization}

\end{table}
\section{Manipulation Experiments}
\begin{figure*}
  \centering
\includegraphics[width=\textwidth]{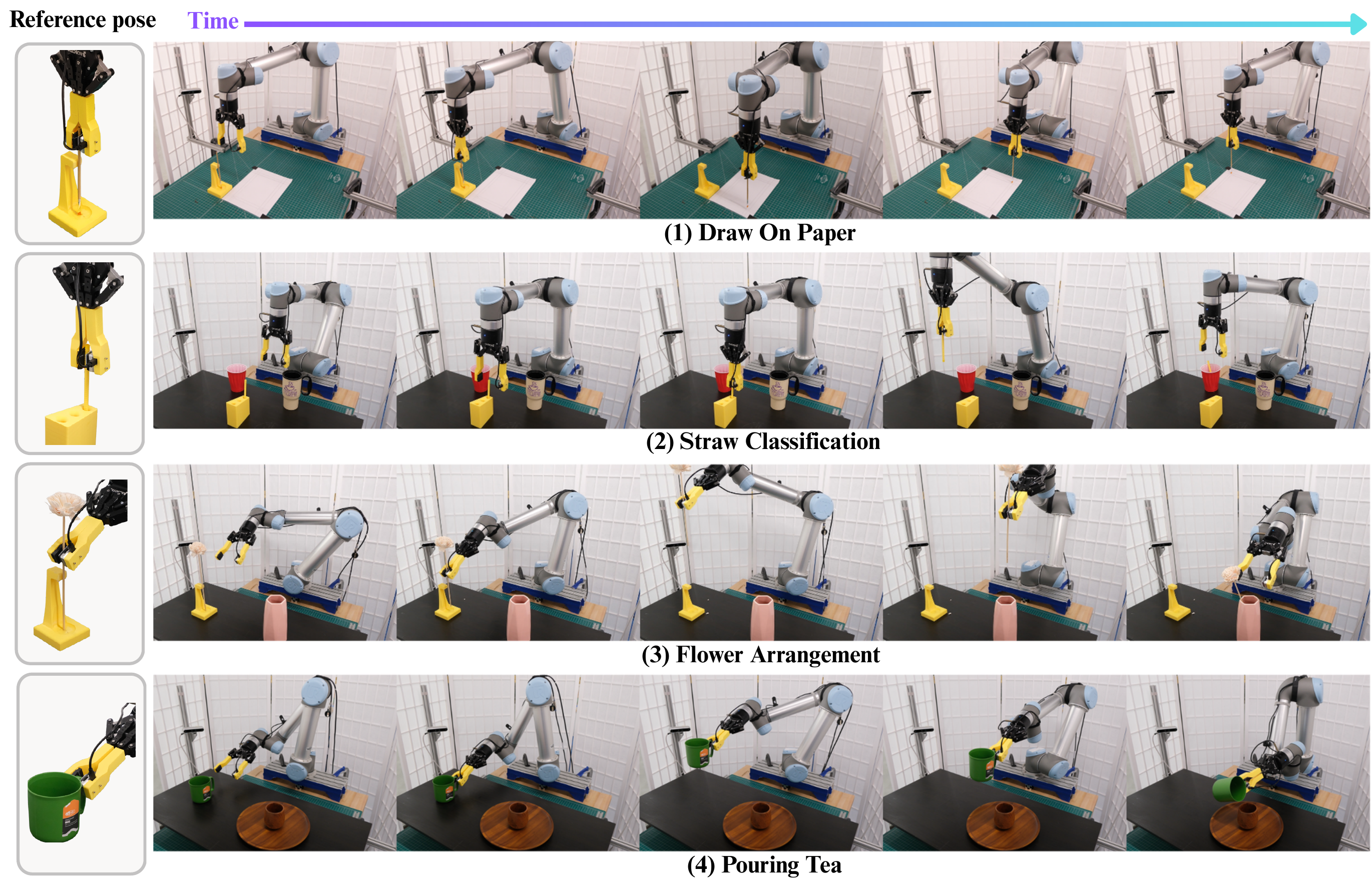}
  \caption{\textbf{Overview of real-world object manipulation experiments.} The left column presents the reference in-hand pose for each object, while the right columns show the robot performing four daily-life tasks including drawing on paper, straw classification, flower arrangement, and pouring tea using tactile feedback to maintain reference in-hand poses.}
  \label{fig:robot_demo}
\end{figure*}

\subsection{Experiment Setup}

In this section, we evaluate our method using a real robot system consisting of a UR5 robot, three Intel RealSense D455 RGBD cameras (RGB channels only), and a GelSight tactile sensor mounted at the center of the gripper, with its center defined as the tool center point (TCP). During data collection, demonstrations are collected using a SpaceMouse controller, with the manipulation object placed in a fixed canonical orientation. During inference, the object is initialized with a random $\SO(2)$ rotation in the gripper’s $z$-plane relative to the canonical orientation, or subjected to external perturbations, to evaluate the model’s robustness to in-hand rotational variations.

We evaluate our correction module across four challenging real-world tasks, as shown in \autoref{fig:robot_demo}.

\begin{enumerate}
\item \textbf{Draw on Paper.}
The robot grasps a hanging paintbrush and draws a rectangle on paper.

\item \textbf{Flower Arrangement.}
The robot grasps a flower stem and inserts it into a vase.

\item \textbf{Straw Classification.}
The robot picks up straws with different surface textures and places them into their corresponding cups.

\item \textbf{Pouring Tea.}
The robot grasps a teacup and performs a pouring motion.

\end{enumerate}

\begin{figure}[t]
    \centering
\includegraphics[width=0.4\textwidth]{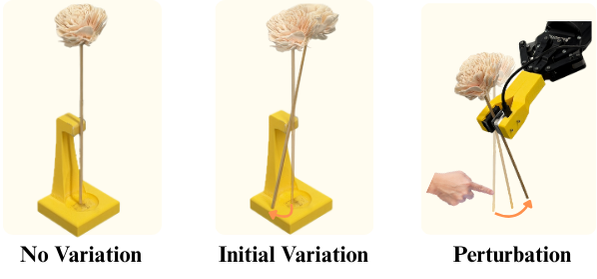}
\caption{\textbf{Illustration of the three evaluation conditions in the Flower Arrangement task.} (a) No Var.: the flower stem starts upright, matching the training distribution; (b) Var. Init.: the stem is initialized with unseen tilted orientations; (c) Pert.: external human disturbance is applied during execution, inducing in-hand orientation shifts.}
    \label{fig:Variation}
\end{figure}

\begin{table*}
\centering   
\caption{\textbf{Real-world Experiment.} Success rate (\%) of 4 physical experiments over 20 evaluation episodes}
\renewcommand{\arraystretch}{1.1}
\setlength{\tabcolsep}{5pt}
\begin{tabular}{l|ccc|ccc|ccc|ccc}
\toprule
\multirow{2}{*}{\textbf{Method}} 
&
\multicolumn{3}{c|}{\textbf{Draw (20 demos)}} &
\multicolumn{3}{c|}{\textbf{Flower (10 demos)}} &
\multicolumn{3}{c|}{\textbf{Straw (40 demos)}} &
\multicolumn{3}{c}{\textbf{Pouring (10 demos)}} \\
\cmidrule(lr){2-13}
& No Var. & Var. Init. & Pert. & No Var. & Var. Init. & Pert. & No Var. & Var. Init. & Pert.& No Var. & Var. Init. & Pert. \\
\midrule
FM (RGB)          & \textbf{1.00} & 0.05 & 0.00 & 0.90 & 0.00 & 0.05 & 0.45 & 0.25&0.40 &0.90 &0.45 &0.40  \\
FM (RGB w/T) & 0.95 & 0.15 & 0.00 & 0.90 & 0.05 & 0.00 & \textbf{0.90} & 0.35& 0.50&\textbf{1.00} &0.30 & 0.35 \\
EquiTac (Ours)           & \textbf{1.00} & \textbf{0.90} & \textbf{0.95} & \textbf{1.00} & \textbf{0.85} & \textbf{0.90} & \textbf{0.90} & \textbf{0.85}& \textbf{0.90}&\textbf{1.00} &\textbf{0.95} &\textbf{0.95}  \\
\bottomrule
\end{tabular}
\label{fig:real_world_data}
\end{table*}

\begin{table}[t]
\centering
\caption{Performance comparison under different tactile information configurations under both \textbf{No Var.} and \textbf{Pert.} conditions.}
\renewcommand{\arraystretch}{1.2}
\setlength{\tabcolsep}{3pt}  
\begin{tabular}{l|cc|cc|cc|cc}
\toprule
\multirow{2}{*}{\textbf{Method}} &
\multicolumn{2}{c|}{\textbf{Draw}} &
\multicolumn{2}{c|}{\textbf{Flower}} &
\multicolumn{2}{c|}{\textbf{Straw}} &
\multicolumn{2}{c}{\textbf{Tea}} \\
\cmidrule(lr){2-9}
 & w/o & w/ & w/o & w/ & w/o & w/ & w/o & w/ \\
\midrule
EquiTac (RGB)  & 1.00 & 0.15 & 0.95 & 0.3 & 0.80 & 0.60 & 1.00 & 0.50 \\
EquiTac (Normal) & 1.00 &0.95  &1.00  &0.80  &0.90  &0.90  &1.00  & 0.95 \\
\bottomrule
\end{tabular}
\label{tab:perturbation_results}
\end{table}

In our experiments, we consider three evaluation conditions as shwon in \autoref{fig:Variation}. 
(1) No Var.: the object is evaluated under the same initial orientation distribution as the training data, without any variation. 
(2) Var. Init.: the object is initialized with unseen orientations, such as the paintbrush or flower starting from a non-vertical pose.
(3) Pert.: external perturbations from human are applied during inference, e.g., the object is poked by a human to alter its in-hand angle. Both Var. Init. and Pert. introduce variations in the in-hand object orientation that are not present during training. The goal is to evaluate whether our correction module enables adaptive robustness under real-world out-of-distribution conditions.
\subsection{Results}

We compare our method against two baselines:

\begin{enumerate}
    \item \textbf{FM (RGB)}: A baseline Flow Matching policy that takes multi-view RGB images and proprioceptive input.
    \item \textbf{FM (RGB w/T)}: A Flow Matching policy that additionally incorporates tactile images alongside multi-view RGB and proprioceptive inputs.
\end{enumerate}

As shown in \autoref{fig:real_world_data}, our method performs similarly as the baseline FM (RGB w/T) under no variations. However, when variations in initial orientation or human perturbations are introduced, 
our method significantly outperforms the baselines. For example, in the \textit{Draw on Paper} and \textit{Flower Arrangement} tasks, where precise contact stability and orientation control are critical, the baseline models (\textit{FM (RGB)} and \textit{FM (RGB w/T)}) show large drops in success rate under perturbations, decreasing from 1.00 to 0.15 and from 1.00 to 0.30, respectively. In contrast, our proposed \textit{EquiTac} maintains a high success rate of 0.95 under both perturbation conditions. This improvement demonstrates that our tactile correction module effectively detects in-hand object pose deviations and generates corresponding compensatory adjustments in real time. 

To better understand the effect of tactile representation, we conduct an ablation study comparing normal-map inputs with raw tactile RGB images in the equivariant correction network. As shown in \autoref{tab:perturbation_results}, replacing the normal map with RGB inputs causes a clear drop in performance across all tasks, especially under perturbation conditions. The reason is that tactile RGB images do not preserve rotation information, making it harder for the network to estimate and correct object orientation accurately. In contrast, normal maps provide direct cues about surface geometry and contact orientation, helping the model maintain stable in-hand poses and recover quickly from external disturbances. These results confirm that combining the equivariant structure with normal-map tactile inputs is key to achieving consistent and reliable correction behavior.

\subsection{Sample Efficiency Evaluation}

\begin{figure}[t]
    \centering
\includegraphics[width=0.47\textwidth]{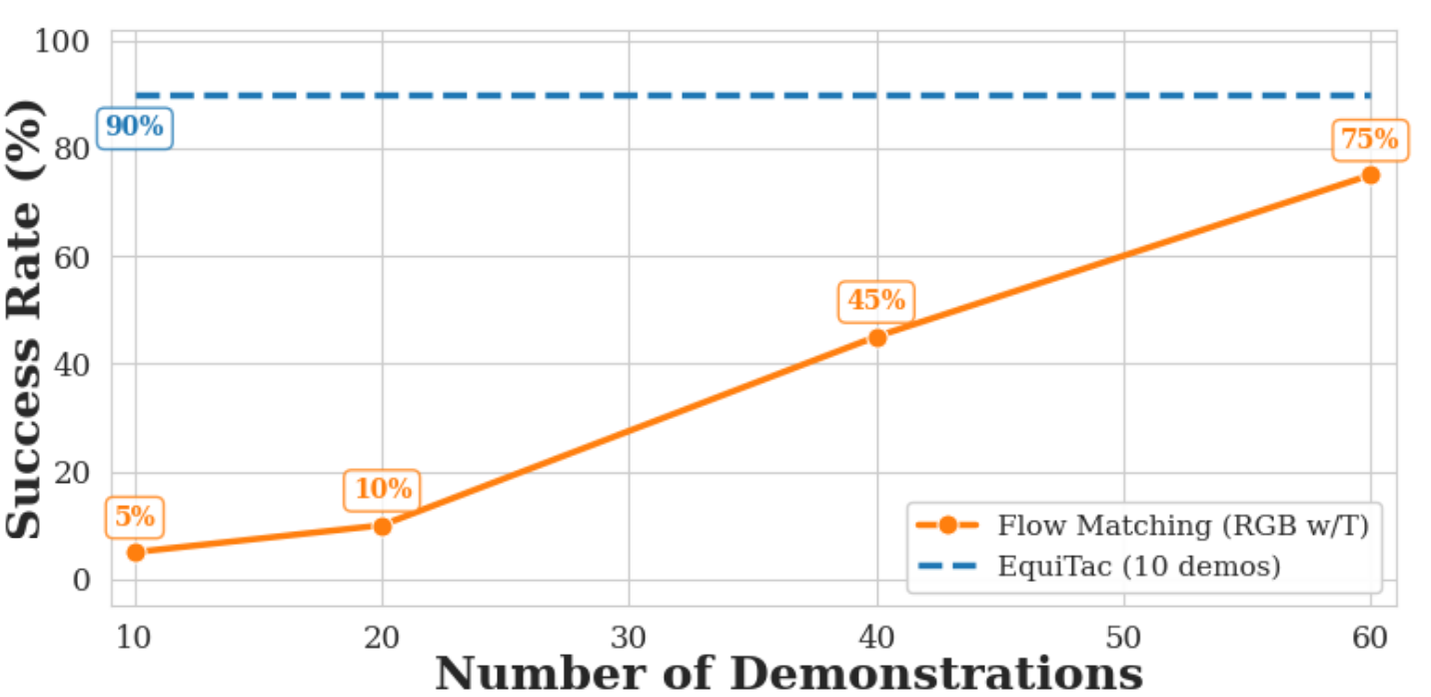}
    \caption{\textbf{Sample efficiency comparison on the Flower Arrangement task.} The Flow Matching (RGB w/T) baseline is trained with 10–60 demonstrations, while EquiTac achieves higher success rates using only 10 demonstrations.}
    \label{fig:Sample_eff}
\end{figure}

To evaluate the sample efficiency of our approach, we augmented the training dataset, which explicitly demonstrates reorientation with variations in the initial object orientation and external perturbations during manipulation. As shown in~\autoref{fig:Sample_eff}, with only 10 demonstrations, \textit{EquiTac} achieves a success rate of 90\%, whereas the baseline reaches at most 75\% even when trained with 60 demonstrations. This result demonstrates that \textit{EquiTac} achieves higher sample efficiency by leveraging equivariant tactile representation, allowing it to learn robust correction behaviors from fewer examples. 

\vspace{5pt}

\section{Conclusion and Limitation}
\textbf{Conclusion.} In this paper, we propose EquiTac, a tactile-equivariant residual correction framework for contact-rich manipulation. By reconstructing surface normal maps and leveraging SO(2)-equivariant representations, our method can precisely estimate the in-hand orientation residual estimation from a single tactile image. We further integrate this equivariant module with a flow-matching visuomotor policy, allowing real-time action correction without additional demonstrations. 

\textbf{Limitations.} Although EquiTac demonstrated excellent performance in practical experiments, the current approach only exploits the rotational symmetry of objects relative to the gripper within the $\mathrm{SO}(2)$ plane. However, in many contact-rich operations, the gripped object may undergo $\mathrm{SE}(3)$ motion relative to the gripper, including rotations and  translations. In addition, collecting multimodal real-world data introduces high cost and limits scalability. In future work, we plan to extend EquiTac to capture $\mathrm{SE}(3)$ changes in tactile feedback and take advantage of tactile simulation to further improve generalization and robustness.


\bibliographystyle{IEEEtran}
\bibliography{reference}

\end{document}